# Local Expression Languages for Probabilistic Dependence: a preliminary report

**Bruce D'Ambrosio**
Department of Computer Science
Oregon State University
(503) 737-5563

## Abstract

We present a generalization of the local expression language used in the Symbolic Probabilistic Inference (SPI) approach to inference in belief nets [1], [8]. The local expression language in SPI is the language in which the dependence of a node on its antecedents is described. The original language represented the dependence as a single monolithic conditional probability distribution. The extended language provides a set of operators (*, +, and −) which can be used to specify methods for combining partial conditional distributions. As one instance of the utility of this extension, we show how this extended language can be used to capture the semantics, representational advantages, and inferential complexity advantages of the "noisy or" relationship.

## 1  Introduction

A belief net [5] is a compact, localized representation of a probabilistic model. The key to its locality is that, given a graphical structure representing the dependencies (and, implicitly, conditional independencies) among a set of variables, the joint probability distribution over that set can be completely described by specifying the appropriate set of marginal and conditional distributions over the nodes involved. When the graph is sparse, this will involve a much smaller set of numbers than the full joint. Equally important, the graphical structure can be used to guide processing to find efficient ways to evaluate queries against the model. For more details, see [5], [7], [1]. All is not as rosy at it might seem, though. The graphical level is not capable of representing all interesting structural information which might simplify representation or inference. The only mechanism available for describing antecedent interactions in typical general purpose belief net inference algorithms is the full conditional distribution across all antecedents. However, a number of restricted interaction models have been identified which have lower space and time complexity than the full conditional. The noisy-or [5], [6], [4] for example, can be used to model independent causes of an event, and is linear in both space and time in the number of antecedents. In this paper we show an extension to the local expression language used in Symbolic Probabilistic Inference (SPI) [8] which is capable of directly expressing a noisy-or interaction model, which captures both the space and time advantages of the model, and which permits use of the model within arbitrary belief nets. In the remainder of this paper we first present a very brief overview of SPI. We then present an extension to the representation used to describe the dependence of a node on its antecedents in SPI, and show how it can be used to capture the noisy-or relationship. We then discuss the nature of the changes which must be made to support this extended local expression language. A key issue is the determination of how to distribute conformal product operations over addition and subtraction. We close with some remaining questions.

## 2  Overview of SPI

In this section we briefly review the essential aspects of the SPI approach to inference in belief nets. For further details, see [1] or [8].

### 2.1  Overview

Computation of probabilities in a belief net can be done quite straightforwardly, albeit somewhat inefficiently[1]. I illustrate this process with a simple network, shown in figure 1. First, the prior probabilities according to the chain rule:

$$p(A) = p(A)$$
$$p(B) = \sum_A p(B|A) * p(A)$$

---

[1] we ignore evidence for purposes of this introduction. It introduces only minor complications, see [1], [8] for details.

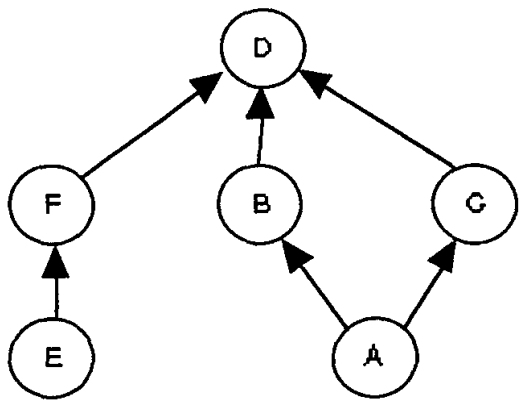

Figure 1: A Simple Belief Net

$p(C) = \sum_A p(C|A) * p(A)$
$p(D) = \sum_{A,B,C,E,F} p(D|B,C,F) * p(F|E)$
$\qquad * p(E) * p(C|A) * p(B|A) * p(A)$
$p(E) = p(E)$
$p(F) = \sum_E p(F|E) * p(E)$

Now suppose we wish to compute $p(D)$. The procedure is quite simple. We begin by evaluating the above expression for $p(D)$ from right to left. Once all distributions are combined, we have computed the joint across all six variables, and can derive the marginal over $D$ by summing over all other variables. The actual computation can be optimized somewhat by retaining each dimension only until we have combined with all terms in which the dimension appears (that dimension is a goal of the evaluation, in which case it must be retained throughout the computation). For example, we can sum over $A$, since it is not needed in the final result, immediately after combining with $p(B|A)$ and $p(C|A)$. Conjunctive queries are easily computed, simply by evaluating the union of the symbolic expressions for the corresponding nodes. SPI essentially follows this process, but can be viewed as a heuristic procedure for developing factorings which minimize the dimension of intermediate results. The factoring is developed incrementally, and factoring is intermixed with expression evaluation. We briefly sketch the process used in the following.

SPI uses a more compact representation, expressing each node marginal only in terms of its conditional distribution and (implicitly) the immediate antecedents needed for computation. For our sample belief net this yields the following expressions:

$$\begin{aligned} exp(A) &= p(A) \\ exp(B) &= p(B|A) \\ exp(C) &= p(C|A) \\ exp(D) &= p(D|BCF) \\ exp(E) &= p(E) \\ exp(F) &= p(F|E) \end{aligned}$$

It may not be obvious how we can reconstruct the earlier computations from this representation. I will describe the evaluation algorithm shortly.

The next component of the representation is a partitioning of the set of nodes. The partitions are arranged in a tree subject to constraints as described in [8], although note that we permit partitions to contain more than one node in this paper. One valid partition of the example belief-net is shown in fig. 2[2].

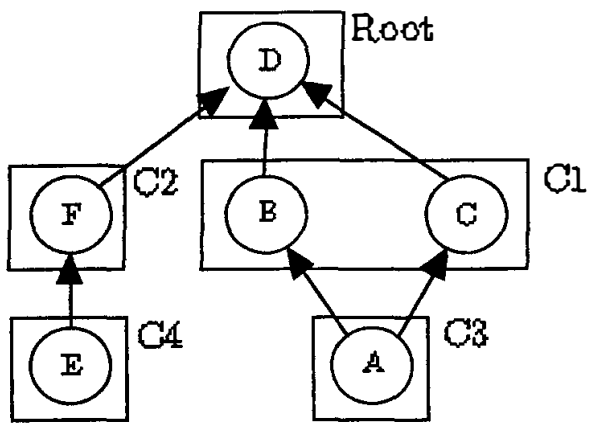

Figure 2: A Partition of the Sample Belief Net

Now consider how we might compute $p(D)$, given this information. The marginal probability of any node can be computed by multiplying the probability expression for the node of interest by distributions across sets of antecedents, subject to the following decomposition constraint: a joint distribution must be computed for antecedents whose corresponding nodes lie in a subtree rooted by the same child partition. Thus, the partitioning describes how to factor any expression at each stage of computation. The expression for $D$ in the root partition, for example, will be decomposed into three groups, $\{p(D|BCF)\}$, $\{p(BC)\}$, and $\{p(F)\}$[3]. $p(D|BCF)$ is known, $p(BC)$ and $p(F)$ will be computed by querying the appropriate child partitions. The identification of the two subqueries that can be processed independently ($p(BC)$ and $p(F)$) is central to the efficiency of SPI. See [8] for proofs of the properties which make this possible.

---

[2]The root of the partition tree need not contain only belief net leaf nodes, and subtrees need not contain only antecedent nodes. We present here what we consider the minimal description of SPI needed to understand the extensions described in this paper.

[3]It should be noted that the value returned from a query to a child partition will be a distribution conditioned on all evidence in the subtree rooted by that child, but not conditioned on evidence elsewhere in the partition tree. Attempting to distinguish the various states of conditioning would clutter the representation, so we will not attempt to indicate the set of evidence a distribution has been conditioned with respect to in this paper.

Since the partition graph is a tree, the recursion will terminate (and be evaluable, since all leaf node marginals are defined in the original belief net).

Below I detail this process for evaluating the marginal probability $p(D)$:

1. In the root partition, determine the expansion for $p(D)$:

$$exp(D) \;=\; p(D|BCF)$$

2. The following child partition queries are formed according to the antecedent set decomposition criterion:
   - Query to C1: $p(BC)$?
   - Query to C2: $p(F)$?

3. C1 expands $p(BC)$ to:

$$p(B|A)p(C|A)$$

4. This in turn generates a query to C3: $p(A)$?

5. C3 returns $p(A)$.

6. C1 evaluates $\sum_A p(B|A)p(C|A)p(A)$ and returns $p(BC)$.

7. C2 expands $p(F)$ to: $p(F|E)$.

8. This in turn generates the query to C4: $p(E)$?

9. C4 returns $p(E)$.

10. C3 evaluates $\sum_E p(F|E)p(E)$ and returns $p(F)$.

11. Root evaluates $\sum_{BCF} p(D|BCF)p(BC)p(F)$ and returns $p(D)$.

This simple example demonstrates a key feature of the algorithm: each partition deals with a low dimensional subspace of the overall probability space. While six variables are involved, the factoring keeps the dimensionality of intermediate computations down to four. We have made several simplifications in this presentation: we consider only partition trees with all belief net root nodes in partition tree leaves, and we do not consider evidence. See the cited papers for a more complete treatment of the basic algorithm.

## 3 Local Expression Languages for Probabilistic Knowledge

In this section we extend the local representation in SPI. This extended expression language is useful for compact representation of a number of canonical interaction models among antecedents. In particular, we demonstrate its use in capturing the noisy-or model. The next section describes the extensions needed in the SPI evaluation algorithm.

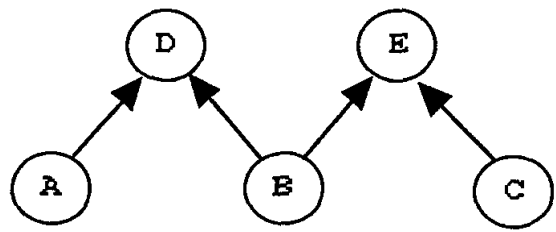

Figure 3: Noisy Or Sample Net

The local expression attached to a node in SPI as described in the previous section is a particularly simple one: it is either a marginal or conditional probability distribution. While this representation is complete (that is, is capable of expressing any coherent probability model), it suffers from both space and time complexity limitations: both the space and time required are exponential in the number of antecedents. However, computation of child probabilities using the noisy-or interaction model is linear in the number of (independent) antecedents in both space and time. When evidence is available on child nodes, computation of the posterior probability of parents is exponential in the number of positive pieces of evidence, but linear in the number of pieces of negative evidence. Heckerman [3] has developed an algorithm, called Quickscore, which provides this efficiency for two level bipartite graphs. However, the author is unaware of any implemented system other than the one reported here which can efficiently incorporate a noisy-or within an arbitrary belief net. If the interaction between the effects of $A$ and $B$ on $D$ in example 3 can be modelled as a noisy-or interaction, then we might write the following expression for the dependence of $D$ on $A$ and $B$, following Pearl [5]:

$$F(D=t) \;=\; 1 - (1 - c_A(D))(1 - c_B(D))$$
$$F(D=f) \;=\; (1 - c_A(D))(1 - c_B(D))$$

Where $c_A(D)$ is the probability that $D$ is true given that $A$ is true and $B$ is false. We use $c$ rather than $p$ to emphasize that these are not standard conditional probabilities. Nonetheless, in the following we will show that we can compute with these distributions[4] using an extension to the same mechanisms already in SPI. There are three components of SPI that must be extended:

1. The expression language must be extended to permit direct encoding of those aspects of interaction

---

[4]As will be noted later, all elements being combined are represented as generalized distributions over some domain. In the case of $c_A(D)$, the domain is the singleton $A = t, D = t$, and $A$ is an antecedent.

models which provide space or time advantages.

2. The symbolic composition operator used to combine local expressions must be extended. This operator previously had one task: to rearrange conformal products of expressions into an efficiently evaluable form, using the commutativity and associativity of the operation. For extended expressions we have distributivity of conformal product over addition and subtraction as an option as well as commutativity and associativity.

3. The numeric evaluation procedure must be extended. First, we must define semantics for the addition and subtraction operators. Second, we must decide how completely to evaluate. The algorithm presented in [8] presumed that it was always appropriate to reduce an expression to the joint distribution over the variables needed in the result. For example, given

$$\sum_B F(D|B,C)F(B)F(C)$$

numeric evaluation will yield $F(D,B)$ rather than $F(D|B)F(B)$. While we believe this to be an optimal choice, the situation is less clear for the extended local expression language. Should the goal of expression evaluation always be to return a joint across the target variables, or are there cases in which it is better to return a partially evaluated expression?

In addition, computational complexity of both symbolic and numeric evaluation stages must remain linear, or at least low order polynomial, in the number of independent antecedents (otherwise what is the point!).

In the following we first present a description of the local expression language we have developed. We then proceed to describe the evaluation of queries referencing nodes whose probabilistic dependency on other nodes is defined using expressions from this language. Not all expressions in this language represent coherent probabilistic relationships. We presume that the user starts with a well understood interaction model and simply needs a computational framework that can perform inference with that model.

### 3.1  BNF for a simple expression language

We present below the BNF for a simple expression language capable of representing noisy-or and a variety of other special-case interaction models:

$$
\begin{aligned}
\textit{arithmetic-exp} \quad &\rightarrow \quad \textit{term} \mid (+\ \textit{term term-set}) \\
&\rightarrow \quad \mid (-\ \textit{term term-set}) \\
&\rightarrow \quad \mid (*\ \textit{term term-set}) \\
\textit{term} \quad &\rightarrow \quad \textit{arithmetic-exp} \mid \textit{distribution}. \\
\textit{term-set} \quad &\rightarrow \quad \textit{term} \mid \textit{term term-set}. \\
\textit{distribution} \quad &\rightarrow \quad \textit{name}_{\textit{dimensions}}. \\
\textit{dimensions} \quad &\rightarrow \quad \textit{conditioned} \text{ ``}|\text{''} \ \textit{conditioning}. \\
\textit{conditioned} \quad &\rightarrow \quad \textit{node-name}_{\textit{domain}} \\
&\rightarrow \quad \textit{node-name}_{\textit{domain}}\textit{conditioned}. \\
\textit{conditioning} \quad &\rightarrow \quad \textit{node-name domain} \\
&\rightarrow \quad \textit{node-name domain} \mid \textit{conditioning}. \\
\textit{domain} \quad &\rightarrow \quad \mid \textit{value} \mid \{\textit{value-set}\}. \\
\textit{value-set} \quad &\rightarrow \quad \textit{value.} \mid \textit{value value-set}.
\end{aligned}
$$

Notice that every term eventually must reduce to one or more *distributions* defined over some domain. As an example, the local expressions for $D$ and $E$ for our sample noisy-or figure are as follows:

$$
\begin{aligned}
exp(D) \ &= \ 1_{D_t} - (1_{D_t} - c_{D_t|A_t}) * (1_{D_t} - c_{D_t|B_t}) \\
&\quad + (1_{D_f} - c_{D_f|A_t}) * (1_{D_f} - c_{D_f|B_t}) \\
exp(E) \ &= \ 1_{E_t} - (1_{E_t} - c_{E_t|B_t}) * (1_{E_t} - c_{E_t|C_t}) \\
&\quad + (1_{E_f} - c_{E_f|B_t}) * (1_{E_f} - c_{E_f|C_t})
\end{aligned}
$$

The above notation may seem a bit obscure. It is perhaps further obscured by the fact that the actual numbers are not represented. The notation $1_{D_t}$ denotes a distribution *named* "1," which is defined over the subspace of the joint probability distribution for the network for which node $D$ holds the value $t$. This distribution contains the single value 1.0 (the actual value is not a necessary consequent of the above notation, but it is convenient to give constants names which correspond to their values.). The expression for $D$, then, can be read as a straightforward recoding of the noisy-or model. It specifies that the distribution for $D$ can be computed as the sum of two components. The first component computes a value for $F(D = t)$, and the second, on the next line, computes the value for $F(D = f)$. Since these two terms are mutually exclusive (as is obvious in this case, since they are defined over disjoint elements of the domain of $D$), they can be combined using simple addition. We specify the following semantics for the operators:

* Conformal product. We assign the same semantics as for standard SPI. When combining distributions defined over differing subspaces of the domain for a node, only those values in the domain for which both distributions are defined need be considered. That is, distributions are implicitly extended with 0.0 in all values for which they are not defined. Thus, $1_{D_T}$ can be seen to specify the distribution $\{1.0, 0.0\}$ over $\{D = t, D = f\}$.

+/- sum/difference. Simple sum or difference of the two terms. This assumes that the terms being combined are mutually exclusive. As before, dis-tributions are extended with zeros for values in the domain over which they are not defined. However, note the following interesting case: how do we compute $1_{D_t} - c_{D_t|A_t}$? The first distribution is defined over $D$, but the second is defined over $D$ conditioned on $A$. Unlike the conformal product case, we cannot directly add or subtract distributions over non-identical sets of variables. We first "normalize" the domains by multiplying *both* by $p(A)$. Suppose $p(A) = \{.1, .9\}$, and $c_{D_t|A_t} = .7$. Then the above computation would yield:

| D/A | $1_{D_t}$ | | $c_{D_t|A_t}$ | | Result | |
|-----|------|------|------|------|------|------|
|     | A=t | A=f | A=t | A=f | A=t | A=f |
| D=t | .1 | .9 | .07 | 0.0 | .03 | .9 |

The above is strictly necessary only if $A$ is needed in the result. If it is not, the alternative is to multiply $c_{D_t|A_t}$ by $p(A)$ and then sum over $A$ before subtracting it from $1_{D_t}$.

Note that $c_{D_t|A_t}$ and $c_{D_f|A_t}$ represent the same noisy-or parameter. Two copies of this parameter are needed in our current expression language to denote its participation in the computation for $D = t$ and in the computation for $D = f$. We have not had time to investigate ways to eliminate this redundancy. There are doubtless other notations that would serve equally well. The above notation is the one used in our implementation, and serves to unambiguously specify the subspace over which each term is defined.

## 3.2 Symbolic composition of extended local expressions

In general, query evaluation in each partition consists of three stages, each of which will require modification. The three stages are:

1. Composition of local expressions for all partition nodes involved in the query.

2. Generation of subqueries to each child partition from which information is needed.

3. Numerical evaluation of the results.

In this section we discuss the first of these points, the symbolic composition, and present an algorithm for distributing conformal product over addition and subtraction. This algorithm yields an efficiently evaluable expression under the following restrictions:

1. There are many ways to construct valid partitioning of nodes. The way we currently use, and which we assume here, is to construct a tree with belief net root nodes at the leaves, and with multiple nodes in a partition where permissible under the partitioning constraints. One such partitioning algorithm is described in [1], and the partition tree constructed by that algorithm is shown in figure 4.

2. Queries to child partitions always return a single joint distribution across query nodes.

We will discuss later research in progress on relaxing these restrictions. The algorithm for distribution of conformal product over addition and subtraction begins with the outermost expression, and is as follows:

1. If the operation in the current expression is a conformal product, then

   (a) group terms with overlapping sets of child partitions from which information is needed.

   (b) distribute conformal product one level down in each group, over those terms in the group either separable into subterms which need antecedents from disjoint subtrees, or which require information from only a subset of the child partition set associated with the group.

   (c) repeat this step on the rewritten expression if the operation is still a conformal product. (can occur when terms are themselves conformal products)

2. Recursively apply step 1 to each term if the current expression is a result of performing a distribution operation.

This procedure assumes that the expressions being combined are initially in efficiently evaluable form, as are $exp(D)$ and $exp(E)$ above.

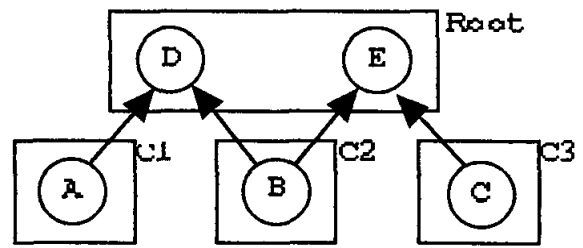

Figure 4: Noisy Or Sample Net Partition

As a simple example, consider the evaluation of the query for $p(D, E)$ in the net shown in Fig.3, where we use the noisy-or model for $\{A, B\}$ and for $\{B, C\}$ (since the queries to child partitions for $p(A)$ through $p(C)$ are trivial, we will ignore them and concentrate on processing in the partitions containing $D$ and $E$). We concentrate on the processing in the root partition.

**Query to root:** $p(D, E)$

**Composition of local expressions for query nodes:**

$$p(D,E) = ((1_{D_t} - (1_{D_t} - c_{D_t|A_t}) * 1_{D_t} - c_{D_t|B_t}))$$
$$+(1_{D_f} - c_{D_f|A_t}) * (1_{D_f} - c_{D_f|B_t}))$$
$$*((1_{E_t} - (1_{E_t} - c_{E_t|B_t}) * (1_{E_t} - c_{E_t|C_t}))$$
$$+(1_{E_f} - c_{E_f|B_t}) * (1_{E_f} * c_{E_f|C_t}))$$

Applying the distribution procedure to the above expression for $p(D,E)$ yields, at the first step (for sake of space we list only the terms involving $D_t$ and $E_t$):

$$p(D_t, E_t) = (1_{D_t} - (1_{D_t} - c_{D_t|A_t}) * (1_{D_t} - c_{D_t|B_t}))$$
$$*(1_{Et} - (1_{E_t} - c_{E_t|B_t}) * (1_{E_t} - c_{E_t|C_t}))$$

Since the top level conformal product in this result contains terms which are separable, it is distributed over those terms. This yields:

$$((1_{D_t} * 1_{E_t}) + ((1_{D_t} - c_{D_t|A_t}) * (1_{D_t} - c_{D_t|B_t}))$$
$$*((1_{E_t} - c_{E_t|B_t}) * (1_{E_t} - c_{E_t|C_t})))$$
$$-((1_{E_t} * ((1_{D_t} - c_{D_t|A_t}) * (1_{D_t} - c_{D_t|B_t})))$$
$$+(1_{D_t} * ((1_{E_t} - c_{E_t|B_t}) * (1_{E_t} - c_{E_t|C_t}))))$$

There are still conformal products which have not been fully distributed. Only one of these, however, contains terms which group together. Distributing that conformal product one level deeper yields:

$$((1_{D_t} * 1_{E_t}) + ((1_{D_t} - c_{D_t|A_t}) * (1_{D_t} - c_{D_t|B_t})$$
$$*(1_{E_t} - c_{E_t|B_t}) * (1_{E_t} - c_{E_t|C_t})))$$
$$-((1_{E_t} * ((1_{D_t} - c_{D_t|A_t}) * (1_{D_t} - c_{D_t|B_t}))$$
$$+(1_{D_t} * ((1_{E_t} - c_{E_t|B_t}) * (1_{Et} - c_{E_t|C_t}))))$$

We are done. We can now apply commutativity and associativity to yield the final evaluation form:

$$((1_{D_t} * 1_{E_t}) + ((1_{D_t} - c_{D_t|A_t})$$
$$*((1_{D_t} - c_{D_t|B_t}) * (1_{E_t} - c_{E_t|B_t}))$$
$$*(1_{E_t} - c_{E_t|C_t})))$$
$$-((1_{E_t} * ((1_{D_t} - c_{D_t|A_t}) * (1_{D_t} - c_{D_t|B_t})))$$
$$+(1_{D_t} * ((1_{E_t} - c_{E_t|B_t}) * (1_{Et} - c_{E_t|C_t}))))$$

None of the conformal products in this final expression meet the criteria for distribution, so we are done, and the expression can be evaluated according to normal SPI methods (the standard SPI local ordering heuristic will group terms appropriately for efficient evaluation, see [1]. The distribution procedure is efficient, and performs well (although not perfectly) at generating an expression which can be efficiently evaluated. We discuss each of these following description of the remaining processing needed.

### 3.3 Subquery Generation:

It should be clear that during the above process the information needed from each child partition has already been identified. Subquery processing proceeds as in standard SPI. It may seem at first that some savings could be achieved by not computing the full distribution for a node, but only the distribution over the referenced subrange. However, at this time we always generate sub-queries for the full distribution across a node.

### 3.4 Expression Evaluation:

We have already discussed the semantics of the operators in the local expression language. The only remaining issue is whether evaluation should always reduce an expression to a single joint distribution across the desired set of result variables or stop short of complete evaluation. Our current implementation always reduces expressions to a single joint distribution.

**Distribution Procedure Complexity** The distribution procedure includes the following steps:

- Grouping of terms - this can be done in $O(nm)$ time, where n is the number of distributions in the expression, and $m$ is the number of child partitions.

- "Separable" test - this can be done in $O(nm)$ time.

- Repetition of these two steps can occur up to $O(l^n)$ times, where $l$ is the length of a node expression, and $n$ is the number of node expressions being composed. For combination of noisy-or expressions as shown above, the overall time will be exponential in the number of expressions being combined which share antecedents.

**Evaluation Complexity** For noisy-or the above procedure preserves the property that the complexity of numeric evaluation is linear in the number of independent antecedents, since independent antecedents will reside in disjoint subtrees below the partition containing the expression being evaluated. As a result, terms referencing them will not group together, and therefore will not force distribution of the conformal product operator. This property is satisfied for the individual noisy or expressions for $D$ and $E$. However, is not true for the initial composition of the local expressions for $D$ and $E$. Correct evaluation of that expression would require that the summation over $B$ be delayed until the subexpressions for $D$ and $E$ had been evaluated and combined. In general, the computation would be exponential in both space and time in the number of shared antecedents. By distributing the conformal product using the algorithm specified above, we reduce the space and time complexity of evaluation of the final expression to linear in the number of shared antecedents. The price we pay, however, is that the expression size, and therefore also evaluation complexity, becomes exponential in the number of nodes being combined. An alternate distribution heuristic might weigh more carefully the costs and benefits of distribution. A few further notes:

1. The "separable" criterion is not perfect. Consider, for example, the expression $((p(D|B,C) +$

$p(E|A,B))*(p(f|A,C)+p(G|B,C)))$ where $A$, $B$, and $C$ reside in disjoint partition subtrees. According to our current test, both terms are separable, since no single distribution requires all of the partitions needed to evaluate an entire term. Nonetheless, distributing the conformal product does not yield a more efficiently evaluable expression.

2. It is not always possible to distribute conformal product operators down to a level which permits independent evaluation of each term (for example, terms might be distributions with overlapping sets of antecedents). In this case the same local evaluation heuristic used in [1] is used to group terms and sequence evaluation.

3. Full distribution of $*$ over $+$ and $-$ would not permit efficient evaluation. Were we to fully distribute conformal product, the result would be correct, but we would need to evaluate a number of terms exponential in both the number of antecedents and the number of nodes.

4. Consideration of the example we presented should make it clear that the algorithm reproduces the essential results of Quickscore when applied to two level bipartite (BN2O) graphs: numeric evaluation is linear in the number of antecedents, linear in the number of negative findings, and exponential in the number of positive findings.

## 4    Discussion

The above procedure is not optimal. It is, however, correct, and therefore provides a method for performing inference using standard interaction models such as noisy-or within SPI. Further, it correctly handles non-independence of antecedents. A review of the example above will reveal that the identification and grouping of the terms involving $B$ did not in any way depend on either the fact that both terms named the same node $(B)$, nor that the terms came from separate local expressions. Similar grouping and distribution of the conformal product operator would have occurred in processing a query for $p(D)$ if $A$ and $B$ were in the same partition subtree below $D$. We have therefore presented a general method for evaluating arbitrary belief nets which contain noisy or models of antecedent interaction.

Noisy-or is traditionally considered to be of restricted applicability since standard presentations restrict to the case where all nodes take only two values. However, there is a straightforward generalization to the multivalued case which requires $(v-1)^2$ parameters for each antecedent, where $v$ is the number of values a variable can take. The methods presented here support this generalization as well as the simple two-value case.

The work presented here is far from complete. Two major extensions are needed to provide efficient support for the local expression language we describe. First, we must extend the distribution heuristic to cover the case where child partitions contain consequent (child) nodes. We believe this to be a minor extension. More difficult is the question of whether it is always appropriate to reduce an expression to a single joint distribution over the query nodes when performing numeric evaluation. In general we have no reason to believe this is the case. The general problem being solved is to find a factoring of the global expression for the query, as described in [1]. The partition tree indicates how to decompose queries and when nodes can be summed over, but contains little further information to guide evaluation. We are therefore investigating techniques which delay expression reduction as long as possible, only performing in each partition the evaluation necessary to perform summing over nodes not needed higher in the tree.

Also, we do not consider the local expression language to be complete. We have begun to explore further extensions to the local expression language. For example, we are pursuing, in conjunction with R. Fung and R. Shachter, the use of a $CASE$ statement to represent contingencies in belief nets [9].

We began our exploration of probabilistic inference in the context of truth maintenance systems, and at that time used symbolic representation at the level of individual probability mass elements [2]. Later, motivated by efficiency concerns, we changed to a symbolic representation at the distribution level [8]. We now seem to have come full circle: the implementation described here again performs symbolic reasoning on elements as small as individual probabilities. The difference is that we now have a choice of representation grain-size, and can select the grain-size appropriate for the dependence model being described.

## 5    Conclusion

Belief nets are a compact, intuitive representation for general probabilistic models, but suffer from inability to efficiently represent low level structural details such as asymmetries and noisy-or relationships. We have shown how the SPI framework can be extended to support a wide class of antecedent interaction models. This permits free use of these models within an arbitrary belief net, and provides efficient processing of arbitrary marginal and conditional queries on the resulting belief net. This facility also provides for easy experimentation on new interaction models, since there is no need to write code to perform inference using the new model: one directly describes the interaction using a simple algebraic local expression language. The full expression language has been implemented and is in use at Intel Corp. in a chip fabrication process di-agnosis project.

## Acknowledgements

Thanks to Bob Fung and Peter Raulefs for many useful discussions. Thanks to NSF (IRI88-21660) for providing the support which made this work possible.